\documentclass[splncs04]{sn-jnl}%

\usepackage{lineno,hyperref}
\usepackage{enumitem}
\usepackage{amsmath,graphicx}
\usepackage{mathtools}
\usepackage{multirow}
\usepackage{hyperref}

\usepackage[normalem]{ulem}
\usepackage{url}
\usepackage{booktabs}
\usepackage{tabularx}
\jyear{2021}%

\theoremstyle{thmstyleone}%
\theoremstyle{thmstyletwo}%

\theoremstyle{thmstylethree}%

\begin{document}

\title[Article Title]{Texture Aware Autoencoder Pre-training And Pairwise Learning Refinement For Improved Iris Recognition}


\author*[1]{\fnm{Manashi} \sur{Chakraborty}}\email{manashi.chakraborty@iitkgp.ac.in}

\author[2]{\fnm{Aritri} \sur{Chakraborty}}\email{aritrichakraborty@gmail.com}

\author[2]{\fnm{Prabir Kumar} \sur{Biswas}}\email{pkb@ece.iitkgp.ac.in}

\author[3]{\fnm{Pabitra} \sur{Mitra}}\email{pabitra@cse.iitkgp.ac.in}

\affil*[1]{\orgdiv{Advanced Technology Development Centre}, \orgname{Indian Institute Of Technology Kharagpur}, \orgaddress{\city{Kharagpur}, \postcode{721302}, \state{West Bengal}, \country{India}}}

\affil*[2]{\orgdiv{Electronics and Electrical Communication}, \orgname{Indian Institute Of Technology Kharagpur}, \orgaddress{\city{Kharagpur}, \postcode{721302}, \state{West Bengal}, \country{India}}}

\affil*[3]{\orgdiv{Computer Science and Engineering}, \orgname{Indian Institute Of Technology Kharagpur}, \orgaddress{\city{Kharagpur}, \postcode{721302}, \state{West Bengal}, \country{India}}}


\abstract{This paper presents a texture aware end-to-end trainable iris recognition system, specifically designed for datasets like iris having limited training data. We build upon our previous stagewise learning framework with certain key optimization and architectural innovations. First, we pretrain a Stage-1 encoder network with an unsupervised autoencoder learning optimized with an additional data relation loss on top of usual reconstruction loss. The data relation loss enables learning better texture representation which is pivotal for a texture rich dataset such as iris. Robustness of Stage-1 feature representation is further enhanced with an auxiliary denoising task. Such pre-training proves beneficial for effectively training deep networks on data constrained iris datasets.
Next, in Stage-2 supervised refinement, we design a pairwise learning architecture for an end-to-end trainable iris recognition system. The pairwise learning includes the task of iris matching inside the training pipeline itself and results in significant improvement in recognition performance compared to usual offline matching. We validate our model across three publicly available iris datasets and the proposed model consistently outperforms both traditional and deep learning baselines for both \textit{`Within-Dataset`} and \textit{`Cross-Dataset`} configurations.
}



\maketitle

\section{Introduction}
In this era of digital impersonation, security of multimedia systems such as smartphones, smart homes, digital personal assistants, banking services, border security etc. becomes extremely important. Iris is one of the most reliable biometric signatures which is not only unique for a person but, as an internal organ (yet patterns are visible externally), iris is resilient to external perturbations and is also stable over time. Thus, iris scanning plays a pivotal role in controlling access grants to such multimedia devices.
\par Traditional approaches of iris recognition involves extracting handcrafted features using predefined filter kernels such as Gabor \cite{daugman1993high,masek2003recognition,ma2003personal}, DCT \cite{monro2007dct}, Laplacian of Gaussian (LOG) \cite{wildes1997iris} etc., which involves empirical filter parameter selection. Most of the traditional methods extract iris descriptors from filter responses of one form of filter kernels, or a combination of few. This limits the capability to represent the complex iris textural patterns. On contrary, a Deep CNN (Convolution Neural Net) is not limited by predefined choice of filter kernels. Instead, a CNN
has the potential to learn data driven filter kernels and are thus more suited for representing complex iris patterns. Invigorated by the enormous success of Deep Learning(DL) in solving image classification \cite{krizhevsky2012imagenet,szegedy2017inception}, object detection \cite{girshick2015fast,redmon2016you}, medical imaging \cite{suk2013deep,li2014deep,lahiri2016deep,lahiri2017generative}, etc., recently iris community has also started to explore the representation prowess of deep nets for iris recognition \cite{nguyen2017iris,gangwar2016deepirisnet,chakraborty2020unsupervised}. Although there is a noteworthy gain of performance in comparison to traditional frameworks, most of the proposed frameworks are inspired from natural images that have distinctive colour and shape information. Iris on the other hand is rich in discriminative textural patterns; however very few works focused into incorporating the domain knowledge of iris while designing the architecture. Also, no attention has been given to cater to small datatset sizes (as low as only 5 samples per subject) in iris recognition. Some of the initial works used transfer learning (using pre-trained models trained on natural images) \cite{nguyen2017iris,minaee2016experimental} to combat dearth of training. However a straight forward transfer learning from natural images having color, shape and structure information to a texture rich domain such as iris might not be a prudent choice.
To the best of our knowledge, our previous work~\cite{chakraborty2020unsupervised} was the first systematic  approach to address these issues with a stagewise learning strategy and incorporating texture aware layers inside a CNN.
\par In this paper we build upon on previous work in \cite{chakraborty2020unsupervised}, both at an architectural and optimization level. For Stage-1 pre-training, we propose a better autoencoding framework with a data relation loss between Gram matrix representations of input and reconstructed images. While curating iris datasets, it is possible to have slight relative changes between head and sensor positions for the same subject during multiple acquisitions. This coupled with sensor noises contribute towards a noisy dataset. In order to learn robust iris features, we further enhance Stage-1 autoencoding with an additional denoising component. Finally, in Stage-2, we initialize a sub part of the network with Stage-1's pre-trained encoder and introduce a pairwise learning strategy which enables end-to-end learning of iris matching. This is in contrast to our previous framework in which iris matching is not part of the training paradigm.

This paper presents the following key technical enhancements over our previous work \cite{chakraborty2020unsupervised}.\\
--~ A better Stage-1 auto-encoding framework with data relation loss combined with usual reconstruction loss; the modified autoencoder network fosters better texture reconstruction which indicates learning better texture aware features \\
--~ Feature learning in Stage-1 with denoising auto-encoding framework: this aids in learning robust iris features\\
--~ Designing a shared network learning strategy for Stage-2 which intakes a pair of Stage-1 features from the corresponding iris pairs and directly returns a matching score; this allows end-to-end learning of iris matcher in contrast to offline matching in our previous work \cite{chakraborty2020unsupervised}. Pairwise learning drives the network to efficiently learn the closeness or discrepancy of an iris pair resulting in appreciable gain in recognition performance compared to our previous offline matching \\
--~ A much more comprehensive evaluation framework with \textit{`all-v/s-all`} matching under a more challenging \textit{`open-world'} configuration across three publicly available iris datasets for both within-dataset and cross-dataset settings. Our approach consistently outperforms both traditional and deep learning based methods across all evaluation settings and datasets.

\section{Related Works} Initially, iris feature extraction was done by extracting hand-engineered features from different filter bank responses. Extraction of features through a pre-defined set of filters limits the widespread representation capability of iris textures. Infact, for such approaches the performance heavily depends on manual parameter selection. In contrast, deep learning  learning paradigms automate the feature learning process with the use of various learnable filter kernels that enhances the feature representation of iris images and usually have better generalization capability. We would first discuss some of the noteworthy traditional approaches and would then move to the recent deep-learning school of thought.
\subsection{Traditional Methods} 
Daugman was one of the pioneers to work in iris recognition \cite{1262028}. Iris features were extracted using 2-D Gabor filters which were binarized to get the \textit{Iriscodes}. Hamming distance between different \textit{Iriscodes} was calculated to evaluate the dissimilarity measure. Masek \cite{masek2003recognition} extracted response from 1D Log Gabor filters. Ma \textit{et al.} \cite{ma2003personal} used multi-scale circularly symmetric sinusoidal modulated Gaussian filters banks to extract iris features. XORing of binarised features gave the dissimilarity score. Wildes  \textit{et al.}~\cite{wildes1997iris} used multiresolution Laplacian of Gaussian (LOG) for extracting representative iris signatures. Matching score was obtained by calculating the normalised correlation between an iris pair. Monro \textit{et al.} \cite{monro2007dct} used features from Discrete Cosine Transform (DCT). To summarize, the traditional works mainly focused on handcrafted iris feature representation. Such pre-defined filter responses of a particular kind limits the iris textures' representation capability. Also, the manual parameter selection of such filter kernels influences the overall recognition efficacy.
\begin{figure*}[!t]
    \centering
    \includegraphics[scale=0.18]{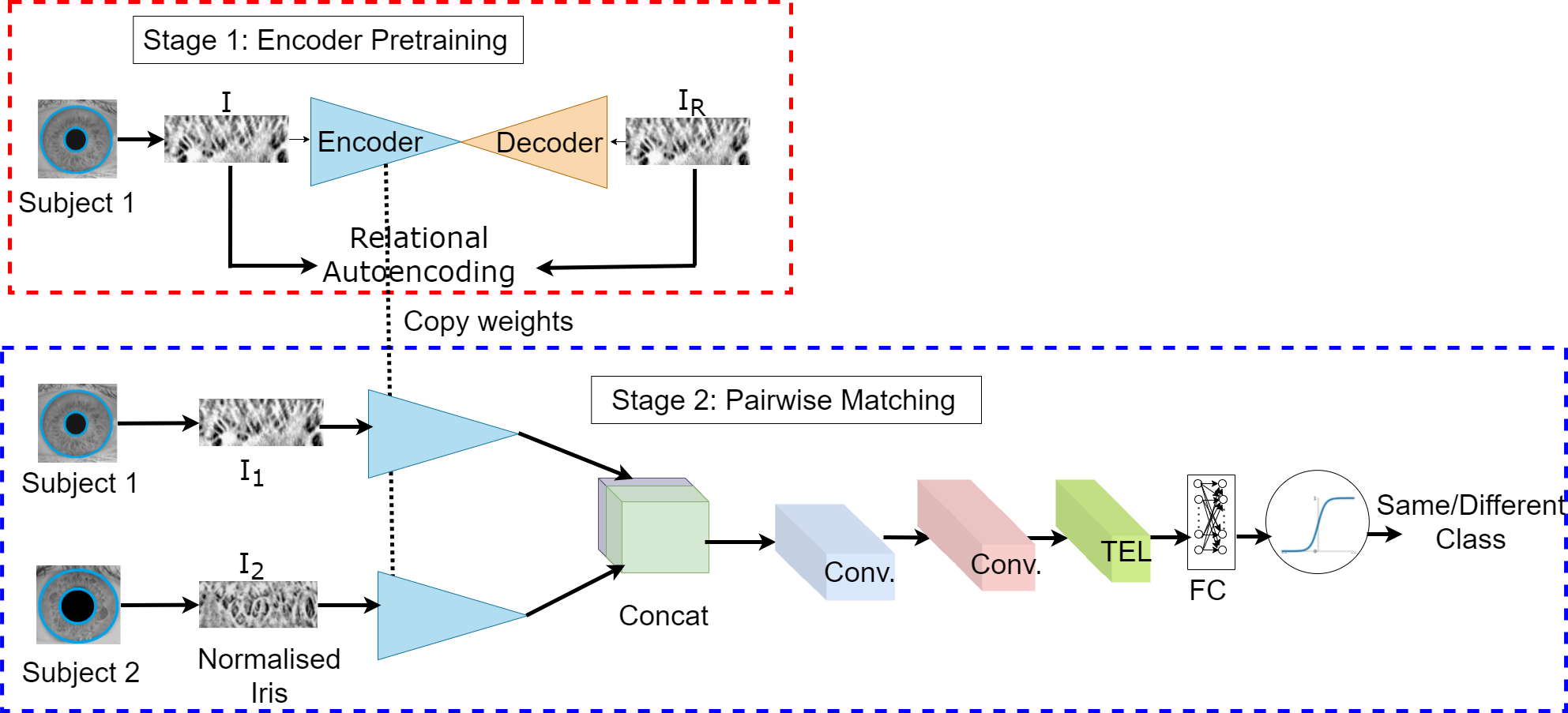}
    \caption{Block diagram of the proposed framework. \textbf{Stage-1:} Unsupervised encoder pre-training using relational autoencoder (Eq. \ref{eq_rel}). \textbf{Stage-2:} Initializing two identical and weight shared copies of encoder with Stage-1 weights followed by supervised fine tuning for learning an end-to-end iris matching using pairwise learning. (Sec. \ref{sec_pairwise_matcher}).}

    \label{fig:workflow}
    
\end{figure*}

\subsection{Deep Learning Approaches} Iris datasets are usually small compared to natural images. Most of the iris datasets have only few thousands of images \cite{phillips2007frvt, kumar2010comparison,casiav4iris} in contrast to natural images that have millions of training data \cite{deng2009imagenet,lin2014microsoft}. Training a complex deep learning model using small train set might overfit the model. Thus, transfer learning was an obvious choice in such scenarios. Minaee \textit{et al.} \cite{minaee2016experimental},Nguyen \textit{et al.} \cite{nguyen2017iris} used kernel weights from popular pre-trained network (for ImageNet classification) previously trained on natural images, refined those weights on iris datasets using supervised classification stage. Minaee et \textit{al.} \cite{minaee2016experimental} for example, refined the features obtained from different layers of VGGNet \cite{simonyan2014very} using SVM classification stage on iris datasets. Menon \textit{et al.} \cite{menon2018iris} used pre-trained ResNet18 and fined tuned it for the iris datasets. However, directly applying transfer learning across such disparate data domain without any special processing might not be apt as natural images have predominance of structure, color or shape information while iris is rich in textural signatures.
\par Gangwar \textit{et al.} ~\cite{gangwar2016deepirisnet} proposed  DeepIrisNet, which used a deep neural network coupled with cross entropy loss to solve the iris classification problem. Once the network is trained a fixed length feature of $4096$ dimensions was extracted for every test image from the second last layer of the trained DeepIrisNet. Euclidean distance between these features are evaluated to calculate the similarity score. Bagar \textit{et al.} \cite{baqar2016deep}  proposed deep belief network based iris classification framework. The architecture is primarily inspired from those used for natural images, they are complex and deep which needs enough training data unlike iris datasets which lacks enormous training samples required for effective training of such complex nets.
\par Our previous work, ICIP'20 \cite{chakraborty2020unsupervised} showcases the benefit of stagewise learning for effective training of data constrained iris datasets. The paper proposed feature learning using unsupervised training in first stage. The second stage or the supervised stage is first initialised with learned weights from stage 1, later fine tuned using classification head. For testing, similar to \cite{gangwar2016deepirisnet}, fixed length $1024D$ features was extracted from the trained network in Stage 2. Euclidean distance between a pair of representative iris features was evaluated to get the dissimilarity score. The stagewise learning concept as stated in this manuscript is inspired from our previous paper, ICIP'20 \cite{chakraborty2020unsupervised}.

\section{Methedology} 
\begin{figure}[!t]
    \centering
    \includegraphics[scale=0.45]{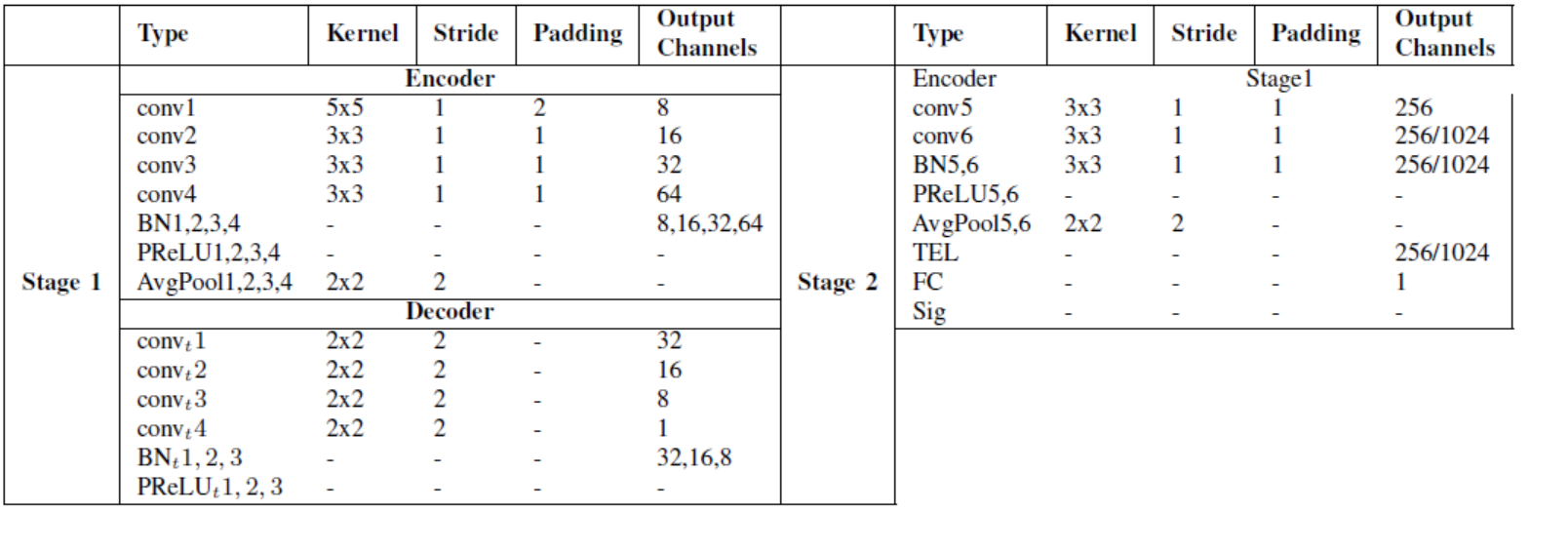}
    \caption{Netwok architecture. \textbf{Stage 1:} Feature pre-training using relational denoising autoencoding. \textbf{Stage 2:} Pairwise matching. Note: BN$i$, PReLU$i$ and AvgPool$i$ are the batch normalization, parametarized ReLU and average pooling used in order mentioned after $i^{ith}$ convolution, conv$i$. $TEL$ is the texture energy layer\cite{chakraborty2020unsupervised}. $FC$ is the fully connected layer and $Sig$ is the sigmoid activation. $BN_ti$ is the batch normalization after transpose convolution, $conv_ti$. For NDIRIS-0405 dataset, output channels of $conv_6$ and $BN_6$ is 1024. For CASIA.v4-Interval and IITD datasets, output channels of $conv_6$ and $BN_6$ is 256.} 
    \label{fig:tab_architecture}
\end{figure}

In this section we elaborate our stagewise training and end-to-end pairwise iris matching framework. We term our composite Stage-1 and Stage-2 framework as \textbf{\textit{CombNet}}. The block diagram is shown in Figure \ref{fig:workflow} and the detailed architecture of both the stages is shown in Figure \ref{fig:tab_architecture}.

\subsection{Stage-1: Autoencoding for Unsupervised Feature Learning} 
In Stage-1, we propose to learn iris encoding using an autoencoder framework trained with unsupervised reconstruction loss and an additional data relation loss \cite{meng2017relational} to learn feature embedding that respects the data relations present in input image. Additionally, we train the autoencoder with a denoising task \cite{vincent2010stacked} to promote learning robust texture features.

\subsubsection{Relational Autoencoder}
A conventional autoencoder first projects a high dimensional data into a low dimensional representation space using an encoder network; then, from this compressed representation a decoder network reconstructs back the input data. The objective function trains the parameters of encoder and decoder such that the reconstruction error is minimised. 
\par If $I$ is the input image, then the reconstructed image, $\hat{I}=  D_{\phi}(E_{\theta}(I))$. Here, $E_{\theta}(\cdot)$ is the encoder (with trainable parameter set $\theta$) and $D_{\phi}(\cdot)$ is the decoder (with trainable parameter set $\phi$). The objective function is,
\begin{equation}
\label{eq_autoen}
   [\theta^*, \phi^*] = \min_{\theta,\phi} L_R(I, D_{\phi}(E_{\theta}(I))) ~.
\end{equation}
Here, $L_R(\cdot)$ is the reconstruction loss.
Information bottleneck theory \cite{tishby2000information} suggests that since we force the encoded space to be of lower dimensional than the input space, the encoded space learns essential feature representation of the input and discards non-relevant information. The encoded features are thus well suited to serve as a feature initialization for the supervised refinement stage, in which, the features can be finetuned based on the final task requirement.

\par One significant drawback of compressing data from the original high dimensional to the encoded low dimensional space without any special care is that it fails to model the inherent data relation as present in original high dimension. Thus, the objective function of conventional stacked autoencoder (Eq.~ \ref{eq_autoen}) that minimises just the reconstruction loss is not enough to capture the complex iris textural patterns/relations. Towards this, we modify the objective function in Eq. \ref{eq_autoen} by incorporating an additional data relation loss, $L_D$. The modified objective function is,

\begin{equation}
\label{eq_rel}
 {[\theta^*, \phi^*]_{Re}} = \alpha \min_{\theta, \phi} L_R(I, \hat{I}) + (1-\alpha)\min_{\theta,\phi} L_D(II^T, \hat{I}\hat{I}^T) ~;
\end{equation}
where $II^T$ is the similarity relation on the original image, $\hat{I}\hat{I}^T$ is the similarity relation on reconstructed image and $\alpha \in [0,1]$ is a relative weight factor. $L_R(\cdot)$ is the usual data reconstruction loss (we used $(1 - SSIM)$) and $L_D(\cdot)$ is the Mean Squared Error (MSE) loss.
\par $II^T$ is the gram matrix representation of input image, $I$. Similarly, $\hat{I}\hat{I}^T$ is the gram matrix representation of the reconstructed image, $\hat{I}$. Gram matrix has been a preferred texture representation block in deep learning based applications such as style transfer \cite{gatys2016image}, texture synthesis \cite{snelgrove2017high}. Gram matrix representation of image $I$ gives a sense of co-relation amongst the image pixels. Minimising the difference between $II^T$  and $\hat{I}\hat{I}^T$ fosters the spatial pixel relation/arrangements of $\hat{I}$ to be similar to input image $I$. Thus, incorporating relational loss along with usual data reconstruction loss is envisioned to drive the Stage-1 network to learn better texture primitives.
We term the Stage-2 network initialised with the encoder optimised with Equation \ref{eq_rel} as $\pmb{CombNet_{E_{\theta}^{REL}}}$.

\subsubsection{Robust feature learning with denoising autoencoder} 
Denoising autoencoder $(DAE)$ \cite{vincent2010stacked} was first proposed to combat the \textit{`identity learning`} \cite{tishby2015deep} problem of stacked autoencoder. The primary objective of DAE is to reconstruct the original input image $I$ from its noisy version, $I_N$. $DAE$ offers the following benefits--  a) learning bottleneck features which are stable and robust against input perturbations; b) denoising task compels the network to learn the data semantics because it has to reconstruct from perturbed pixels. Both of these features are practically important for iris biometrics because during image acquisition there can be various sources of errors such as sensor noise, shift in relative sensor-head positioning, etc. 
In this work, we combine the benefits of data relation preservation of relational autoencoder and robust feature learning of denoising autoencoder into a single combined objective as follows:
\begin{equation}
\label{eq_de_rel}
\begin{split}
[\theta^*, \phi^*]_{Re+De} = \alpha \min_{\theta,\phi} L_R(I, \hat{I_N}) \\ 
+ (1-\alpha)\min_{\theta, \phi} L_D(II^T, \hat{I_N} \hat{I_N}^T)
\end{split}
\end{equation}
where $\hat{I_N}$ is reconstruction of original image $I$ but from it's  noisy input version, $I_N$. In this work we get $I_N$ by adding noise sampled from a Normal distribution;

\begin{equation}
    I_N = I + \mathcal{N} (0, \sigma)~,
    \label{eq_corrupted_input}
\end{equation} 
where $\sigma$ is the standard deviation. We term this variant of proposed network as $\pmb{CombNet_{E_{\theta}^{REL+De}}}$.
\subsection{Stage-2: Pairwise Matcher}
\label{sec_pairwise_matcher}
This is the supervised pairwise matching stage. Unlike offline matcher \cite{chakraborty2020unsupervised,gangwar2016deepirisnet} where the model training pipeline is agnostic of end task of matching, our Stage-2 pairwise matcher network incorporates the end task of matching within the training pipeline. Thus, in this stage, objective of the training pipeline is to learn whether a given iris pair is from same/different class.

\par This stage is initialised with weights of Stage-1 encoder, $E_{\theta}(\cdot)$ which has been optimised using Eq. \ref{eq_de_rel}. In this stage the encoder will be fine-tuned using supervised shared network learning strategy shown in Figure \ref{fig:workflow}. We abbreviate this \textbf{E}nd-to-\textbf{E}nd pairwise \textbf{M}atching network as $\pmb{E^2M\-CombNet_{E_{\theta}^{REL+De}}}$.  As shown in Figure \ref{fig:workflow}, two identical copies (weights shared) of Stage-2 encoders are first initialised with the weights of Stage-1. The encoding section is proceeded by the pairwise matching sub network which is randomly initialized. The entire Stage-2 network is then refined using supervised matching loss. 

\par Specifically, let $E_{\theta}(I_i^c)$ is the iris feature representation from the encoder output for an $i^{th}$ training image from class $c$. Similarly,
we can have another representation as $E_{\theta}(I_j^k)$. The objective of pairwise matcher is to determine if $E_{\theta}(I_i^c)$ and $E_{\theta}(I_j^c)$ belong to same class or not. This stage is optimized using binary cross entropy loss between the target label, $y$ and predicted score from the network. The target, $y$ is defined as 

\[
    y = 
\begin{dcases}
    0,& \text{if }  c = k~~ (i.e., ~~same~~ class)\\ 
    1, & \text{if } c \neq k ~~ (i.e., ~~different ~~class)
\end{dcases}
\]

\section{Dataset Description } \label{dataset}
In this section, we describe the different datasets used for our experiments. 
\subsection{Benchmark Datasets} 
We have performed our experiments on three popular iris datasets, namely, \textit{ND-Iris-0405} \cite{phillips2007frvt}, \textit{CASIA.v4-Interval} \cite{casiav4iris} and \textit{IITD} \cite{kumar2010comparison}. For iris recognition, left and right iris of same person are considered to be two different classes \cite{gangwar2016deepirisnet,chakraborty2020unsupervised}. So, effectively, number of classes is always twice the number of subjects.The details of the datasets are as follows:

\begin{enumerate}
    \item \textbf{ND-Iris-0405:} This is one of the largest publicly available iris dataset. There are a total of 64,980 images from 356 subjects (712 classes). From each subject, first 25 left iris images are taken for training and first 10 right iris images are used for testing. For subjects with less number of images than the aforementioned numbers, all left iris images are used for training and all right iris images are used for testing. 
    \item \textbf{CASIA.v4-Interval:} This dataset is a subset of the CASIA.v4-Iris dataset. There are a total of 2,639 images from 249 subjects (498 classes). From each subject, all the left iris images are taken for training (on average 8 images/subject). First 10 right iris images are used for testing. For subjects with less number of images than the aforementioned numbers, all right images are used for testing. 
    \item \textbf{IITD Iris Database} This database comprises of 2240 images from 224 subjects (448 classes). All left iris images of all subjects form the training set (on average 5 images/subject) and all the right iris images comprise the test set (on average 4 images/subject). 
    
\end{enumerate}

\subsection{Iris pre-processing}
The iris images are first segmented and then normalized (transformation of annular segmented iris from \textit{cartesian} to \textit{polar} coordinate to get rectangular patch) using OSIRIS v4.1 \cite{OSIRIS}. For fair comparison, we follow the same pre-processing steps across all datasets and for all competing methods.

\section{Experimental Setup}
\subsection{Training Protocols}  
All the models are trained with 32 batch size for about 8 hours on NVIDIA GTX 1080Ti card with 11GB memory. The models are trained using \textit{Adam} \cite{kingma2014adam} optimizer with learning rate 0.001 using \textit{Pytorch 1.0} library \cite{paszke2019pytorch}. Iris images of resolution $64\times512$  are used for all  methods except for VGG-16, for which the image resolution is $224\times224$. In Eq. \ref{eq_corrupted_input}, $\sigma$ is kept at $0.25$. For pairwise learning, to maintain homogeneity, a given batch has equal proportions of genuine and imposter pairs. Such pairs are selected randomly.
\subsection{Testing Protocols} \label{configuration}
\subsubsection{Open world all-v/s-all testing framework}
We evaluate recognition performance under the challenging \textit{`open world`} setting in which the test set class identities have neither been used during training nor have been enrolled in the system. We perform an exhaustive \textit{`all-v/s-all`} matching wherein matching score of each possible iris pair is considered for testing. If a test example is matched with another test example of same class, it accounts for a True Acceptance (TA) else it accounts for a False Reject (FR). If a test example is matched with an example from another class, it accounts for False Acceptance (FA).
\subsubsection{Testing Configurations} \label{test_config}
Performance of various methods are evaluated under the following two testing configurations:
\begin{enumerate}
    \item \textbf{Within Dataset:} In this configuration, training and test data are partitioned from the same dataset. In this setting even though test time class identities are mutually exclusive of training class identities, the image capture setup are same for train and test set as both the partitions are selected from the same dataset.
    \item \textbf{Cross Dataset:} In this setting, the model is trained on one dataset and tested on a completely different dataset without any further fine-tuning on the test dataset. This is more challenging because image capture settings are different across datasets. Cross dataset evaluation usually speaks of the robustness and generalization capability of a iris recognition framework and also mimics a product deployment scenario.
\end{enumerate}
\subsection{Evaluation Metrics}
To benchmark performances of different competing methods, we report the \textit {Equal Error Rate (EER)}  and \textit{Area under Curve (AUC)} $\in [0, 1]$ of the Detection Error Tradeoff (DET) curve. EER is the point where difference between \textit{FAR (False Acceptance Rate)} and \textit{FRR (False Rejection Rate)} is minimum. For a good recognition system, smaller value of EER is preferred \cite{chakraborty2020unsupervised, gangwar2016deepirisnet}. DET curve is the plot of FRR versus FAR. For a good recognition system, it is desirable to have a low value for AUC under DET curve.  We also plot DET curves to visually compare performances of different competing methods.
 

\subsection{Comparing Methods}
Our proposed model is compared against various traditional as well as recent deep learning frameworks. The methods are listed as below: 

\begin{enumerate}
    \item \textbf{Traditional Methods:} We compared against Daugman \cite{1262028}, Masek\cite{masek2003recognition} and Ma \textit{et al.} \cite{ma2003personal}. Feature extraction of all the traditional methods were implemented using USIT (University Of Salzburg Iris Toolkit) \cite{USIT3}. Traditional methods can only be compared for \textit{`Within-Dataset`} configuration.
     In \textit{`Cross-Dataset`}  configuration, a model is trained on one dataset and tested on a different dataset. As traditional methods do not involve any learning stage, \textit{`Cross-Dataset`} configuration is not applicable for these methods.
    \item \textbf{Deep-Learning Approaches:} We have compared our method against the typical DL techniques used for iris recognition. They are listed below:
    \begin{enumerate}
        \item \textit{Transfer Learning:} 
        An obvious deep learning baseline on a data constrained domain such as iris is to finetune any popular ImageNet pre-trained networks. Here, we finetuned a pre-trained \textbf{VGG-16}\cite{simonyan2014very}. Before finetuning, we replace the last layer of VGG-16 to match the number of classes  on respective datasets.
        \item \textit{ Training Deep Conventional CNN Frameworks From Scratch:} We have compared our work with \textbf{DeepIrisNet} \cite{gangwar2016deepirisnet} which is a direct application of conventional CNN frameworks. Unlike transfer learning, DeepIrisNet trains the entire model from scratch.
        \item \textit{ Stagewise Training:} We have also compared against our previous work, \textbf{ICIP'20}. \cite{chakraborty2020unsupervised}.
    \end{enumerate}
    We have evaluated the performances of above methods on both \textit{`Within-Dataset`} and \textit{`Cross-Dataset`} configurations. It is to be noted that except for our proposed method, all the remaining deep learning models process a single image at a time and use offline matcher \cite{chakraborty2020unsupervised, gangwar2016deepirisnet} based on features tapped from their respective tapping points. For ablation of offline-matching variants of our architectural choices (refer to Table \ref{tab_ablation}), $CombNet_{R}$, $CombNet_{E_{\theta}^{REL}}$ and $CombNet_{E_{\theta}^{REL+De}}$, features are tapped from the penultimate fully connected layer of the architecture shown in Figure \ref{fig:tab_architecture}. The offline matcher module of all the competing DL methods are adapted from \cite{chakraborty2020unsupervised, gangwar2016deepirisnet}, and therefore euclidean distance between a given iris pairs gives the dissimilarity score.
\end{enumerate}
\begin{table}[]
\centering
\caption{ Ablation study of variants of architectural choices. ($\downarrow$) is better.}
\label{tab_ablation}
\begin{tabular}{|l|l|l|l|l|ll}
\cline{1-5}
Model                & EER (in\%) $\downarrow$ & AUC $\downarrow$ & Offline Matcher? & Pre-trained Encoder? &&\\ \cline{1-5}
$CombNet_{R}$           & 6.08           &  0.018           &  Yes   & No &  \\
$ICIP'20$ \cite{chakraborty2020unsupervised}          &  5.01          &   0.015         &  Yes  & Yes &  \\
$CombNet_{E_{\theta}^{REL}}$ & 4.81           & 0.013         &  Yes    &  Yes &  \\
$CombNet_{E_{\theta}^{REL+De}}$     &  4.72          &0.011         &  Yes     & Yes &  \\ 
$E^2MCombNet_{E_{\theta}^{REL+De}}$       &  \textbf{1.78}         & \textbf{0.002}         &  No    & Yes &  \\\cline{1-5}
\end{tabular}
\end{table}
\section{Results}
We first present an ablation analysis of different variants of our architectural choices followed by comparative analysis against competing methods.
\subsection{Ablation Study}
Here we study the efficacy of variants of our architectural choices on NDIRIS-0405 under \textit{Within Dataset} configuration (Section \ref{test_config}). 
\subsubsection{\textbf{Pre-training of Encoder versus Randomly Initialized Encoder}} Here we first show the benefit of encoder pre-training (via autoencoder training) over a randomly initialised encoder, $CombNet_R$. We compare our auto-encoder pre-trained network, $CombNet_{E_{\theta}^{REL}}$ against randomly initialized $CombNet_{R}$.
\par $CombNet_R$ achieves (6.08, 0.018) for (EER, AUC), while $CombNet_{E_{\theta}^{REL}}$ achieves (4.81, 0.013) -- an improvement of $20.88$\% for EER and $27.77$ \% for AUC.
Recognition performance is further enhanced with our additional design choices of incorporating denoising auto-encoding and pairwise learning (to be discussed in next sections).
From Table \ref{tab_ablation}, the trend of superiority in performance can be observed for models with pre-trained encoder over randomly initialised encoder.

\par Such improvement of performance with stagewise learning is in line with the observation of our previous work, ICIP'20 \cite{chakraborty2020unsupervised}. In this paper, we enhanced the performances of ICIP'20 further with our proposed design choices.
\subsubsection{\textbf{Benefit of Encoder Pretraining with Relational Autoencoder}} Here we show the benefit of using relational autoencoder over conventional autoencoder as used in ICIP'20 \cite{chakraborty2020unsupervised} for encoder pre-training. $CombNet_{E_{\theta}^{REL}}$ is the Stage-2 recognition framework whose encoder is pre-trained in Stage-1 using relational loss (Eq. \ref{eq_rel}).  From Table \ref{tab_ablation} we can see
that $CombNet_{E_{\theta}^{REL}}$ achieves a relative improvement of 4\% and 13.33\% in terms of EER and AUC compared to ICIP'20.
\begin{figure}[!t]
    \includegraphics[scale=0.6]{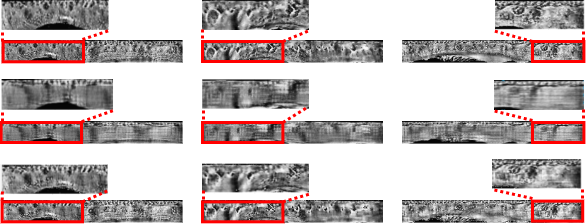}
    \caption{Visual comparison of texture reconstruction of proposed versus our previous work, ICIP'20 \cite{chakraborty2020unsupervised} on three different iris images. \textbf{Top Set:} Original images, \textbf{Middle Set:} Reconstruction from ICIP'20, \textbf{Bottom Set:} Reconstruction from \textbf{proposed} method. It can be seen clearly that proposed method fosters better reconstruction of texture patterns which is a signature of learning better texture primitives. Best viewed when zoomed in.}
    \label{fig:IITD_texture_reconstruct}
    
\end{figure}

         
\par We attribute the better performance of $CombNet_{E_{\theta}^{REL}}$ to a better texture representation learning during encoder pre-training in Stage-1. First, an encoder learns to represent an iris image as a compact texture representation. Second, we observe that the texture reconstruction of $CombNet_{E_{\theta}^{REL}}$ is superior to that of \cite{chakraborty2020unsupervised}. In Figure \ref{fig:IITD_texture_reconstruct}, we compare some example reconstructions from Stage-1 of $CombNet_{E_{\theta}^{REL}}$ and ICIP'20 on three different iris images. $CombNet_{E_{\theta}^{REL}}$ is able to reconstruct finer texture details compared to ICIP'20. Addition of the relational loss thus encourages the network to learn textures aware features which fosters better texture reconstruction.
Since Stage-2 finetuning is common for both of the methods, the enhanced recognition performance can be attributed to a better encoder pre-training in Stage-1 of $CombNet_{E_{\theta}^{REL}}$ enabled by the additional data relation loss.\\
         
\subsubsection{\textbf{Benefit of Encoder Pretraining with Relational Denoising Autoencoder}}
Here we show the benefit of augmenting additional denoising loss (Eq. \ref{eq_de_rel}) while training the relational autoencoder. We term this combined architecture as $CombNet_{E_{\theta}^{REL+De}}$. From Table \ref{tab_ablation}, it is observed that $CombNet_{E_{\theta}^{REL+De}}$ achieves a relative gain of 1.90\% and  15.38\% on EER and AUC compared to $CombNet_{E_{\theta}^{REL}}$. 
Corresponding gains with respect to ICIP'20 are 5.8\% and 26.4\%.
Iris images can be perturbed by different noise sources such as sensor noise, sensor-head misalignment etc.  Denoising autoencoding aids extraction of features resilient against such perturbations which manifests in better recognition performance.
         
\subsubsection{\textbf{End-to-End Pairwise  Matching versus Offline Matching}} Here we show the benefit of learning an end-to-end trainable iris pair matcher over the usual paradigm of offline matching. Usually an iris recognition has a feature learning stage, followed by an offline matcher module that computes the matching score between the iris features through some deterministic distance measure. In contrast, our pairwise matcher framework, $E^2M\-CombNet_{E_{\theta}^{REL+De}}$, directly predicts if an iris pair belongs to same/different class. Thus, end task of matching is incorporated into the training pipeline.
          
As observed from Table \ref{tab_ablation}, $E^2M\-CombNet_{E_{\theta}^{REL+De}}$ gives the best performance with an EER of 1.78 and AUC of 0.002. This is a significant improvement of 62.28\% on EER and 81.81\% on AUC over our best offline matching framework of $CombNet_{E_{\theta}^{REL+De}}$ while the improvements over our ICIP'20 \cite{chakraborty2020unsupervised} are 64.47\% and 86.66\% respectively. The observations strongly encourages end-to-end training of pairwise iris matching instead of the usual practise of offline matching.
\par We hypothesize that the pairwise learning strategy promotes better learning of differential features from a pair of images and ultimately fosters in better end result of matching task.
\par Going forward, based on the observations of ablation study, we always refer to \textbf{$E^2M\-CombNet_{E_{\theta}^{REL+De}}$} as our \textbf{proposed} method, unless otherwise stated.
\begin{figure}[!t]
    \includegraphics[scale=0.3]{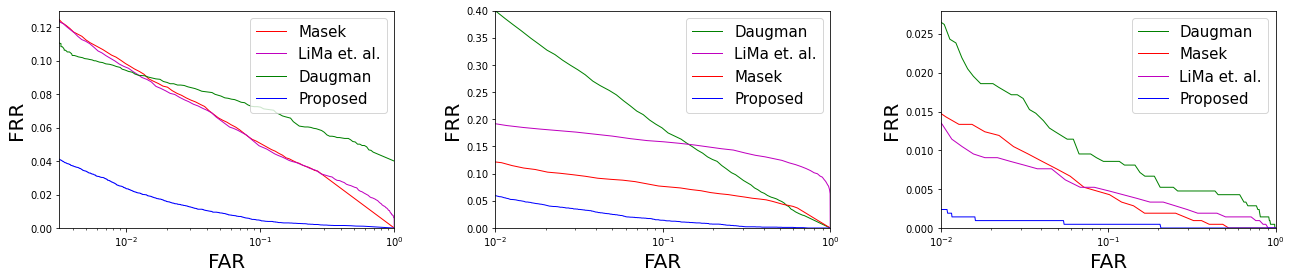}
    \caption{Comparative DET curves of proposed method and three popular traditional baselines on: \textbf{Left:} NDIRIS-0405, \textbf{Middle:} CASIA.v4-Interval, \textbf{Right:} IITD  datasets under \textit{Within Dataset} matching configuration. The curve closest to the bottom is desirable for better recognition performance, which is the case for our proposed model.}
    \label{fig:traditional_DET}
    
\end{figure}


\begin{table}[]

\centering
\caption{ Benchmarking the performance of our proposed model against traditional methods on \textbf{NDIRIS-0405} dataset under 
\textit{Within Dataset} testing configuration. ($\downarrow$): lower is better.}
\label{WDB_trad_ndiris}
\begin{tabular}{|l|l|l|ll}
\cline{1-3}
Model                & EER (\%)~$\downarrow$ & AUC~$\downarrow$ &  &  \\ \cline{1-3}
Daugman \cite{1262028}       & 7.56           &  0.053   &  &  \\
Masek \cite{masek2003recognition}          &  6.11          &   0.026  &  &  \\
Li Ma et \textit{al.} \cite{ma2003personal} & 5.96           & 0.028    &  &  \\

Proposed     &  1.78         & 0.002     &  &  \\\cline{1-3}
\end{tabular}
\end{table}
\begin{table}[]

\centering
\caption{ Benchmarking the performance of our proposed model against traditional methods on \textbf{CASIA.v4-Interval} dataset under 
\textit{Within Dataset} testing configuration. ($\downarrow$) is better.}
\label{WDB_trad_Interval}
\begin{tabular}{|l|l|l|ll}
\cline{1-3}
Model                & EER (in\%) $\downarrow$ & AUC $\downarrow$ &  &  \\ \cline{1-3}
Daugman \cite{1262028}       & 14.97          &  0.077   &  &  \\
Masek \cite{masek2003recognition}          &  8.05          &   0.012  &  &  \\
Li Ma et \textit{al.} \cite{ma2003personal} & 15.33           & 0.047    &  &  \\

Proposed     &  3.33         & 0.005    &  &  \\\cline{1-3}
\end{tabular}
\end{table}
\begin{table}[]

\centering
\caption{ Benchmarking the performance of our proposed model against traditional methods on \textbf{IITD} dataset under 
\textit{Within Dataset} testing configuration. ($\downarrow$) is better.}
\label{WDB_trad_IITD}
\begin{tabular}{|l|l|l|ll}
\cline{1-3}
Model                & EER (in\%) $\downarrow$ & AUC $\downarrow$ &  &  \\ \cline{1-3}
Daugman \cite{1262028}       & 1.87          &  0.0060   &  &  \\
Masek \cite{masek2003recognition}          &  1.29          &   0.0030  &  &  \\
Li Ma et \textit{al.} \cite{ma2003personal} & 1.16           & 0.0040    &  &  \\

Proposed     &  0.85         & 0.0001    &  &  \\\cline{1-3}
\end{tabular}
\end{table}

\subsection{Comparison with Existing Methods} We compare the performance of our final proposed model ($E^2M\-CombNet_{E_{\theta}^{REL+De}}$) against various traditional as well as deep learning frameworks across \textit{Within Dataset} and \textit{Cross Dataset} testing configurations.\\
\textbf{Within Dataset Performance:}  
Here we compare efficacy of our proposed model against both traditional and deep learning approaches on NDIRIS-0405, IITD and CASIA.v4-Interval datasets.

\par Tables \ref{WDB_trad_ndiris}, \ref{WDB_trad_Interval} and \ref{WDB_trad_IITD} shows the performances of our proposed model against three popular traditional baseline architectures on NDIRIS-0405, CASIA.v4-Interval and IITD datasets respectively. Relative gain of our proposed method in terms of EER over Daugman, Masek and  Li Ma et. \textit{al.} are respectively 76.45\%, 70.86\%, 70.13\% on NDIRIS-0405. On CASIA.v4-Interval the respective gains are 77.75\%, 58.63\% and 78.27\% and on IITD the respective gains are 54.54\%, 34.10\% and 26.72\%. In terms of AUC, the relative gain over Daugman, Masek and  Li Ma et. \textit{al.} on NDIRIS-0405 are 96.22\%, 92.30\% and 92.85\% respectively. On CASIA.v4-Interval the respective gains are 93.50\%, 92.30\% and 92.85\% respectively. On IITD the respective gains are 98.33\%, 96.66\% and 97.50\% respectively. Although our proposed method outperformed the traditional baselines on all the three datasets, the relative gain of proposed framework is less in IITD dataset in comparison to the gains on the other two datasets.
This can be attributed  to the very small training dataset size of IITD. Traditional techniques extract features from handcrafted filters with deterministic filter parameters and thus do not have training stage and therefore do not require huge volumes of data for training.
Deep learning architectures on the contrary requires huge volume of data for fruitful training \cite{simonyan2014very,krizhevsky2012imagenet}. However, our texture attentive stagewise architectural choices coupled with pairwise learning scheme, efficiently combats the prominent scarcity of training data in IITD dataset and still manages to outperform the traditional baselines. Figure \ref{fig:traditional_DET} compares the DET curves which again bolsters the superior recognition performance of our model across all the three datasets.
\begin{figure}
    
    \includegraphics[scale=0.3]{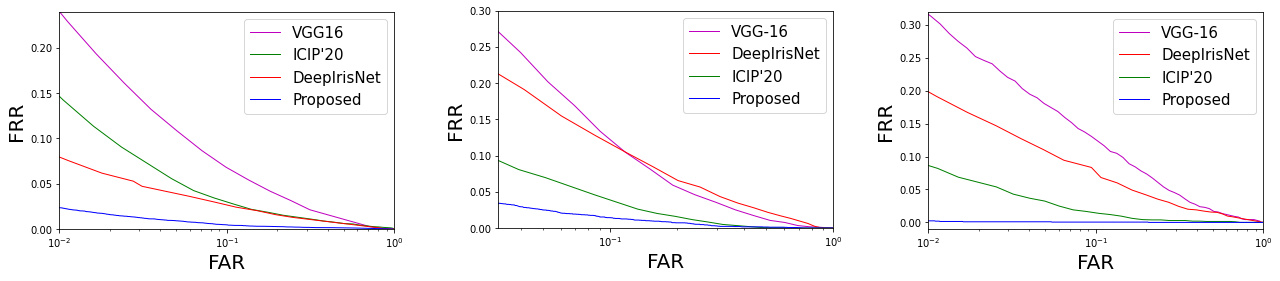}
    \caption{Comparative DET curves of proposed method against deep learning baselines on: \textbf{Left:} NDIRIS-0405, \textbf{Middle:} CASIA.v4-Interval, \textbf{Right:} IITD datasets under \textit{Within Dataset} matching configuration. The curve closest to the bottom is desirable for better recognition performance, which is the case for our proposed model.}
    \label{fig:Deep_DET}

\end{figure}

\par Tables \ref{WDB_deep_Ndiris}, \ref{WDB_deep_Interval} and \ref{WDB_deep_IITD} compares the performance of our proposed method against recent deep learning techniques on NDIRIS-0405, CASIA.v4-Interval and IITD datasets respectively. On NDIRIS-0405, our method outperforms finetuned VGG-16 by 77.38\%, DeepIrisNet  by 64.75\% and ICIP'20 by 64.25\% in terms of EER. On CASIA.v4-Interval the corresponding gains are 69.94\%, 68.73\% and 45.49\%. On IITD dataset, the respective gains are 92.41\%, 89.26\% and 77.86\%. In terms of AUC, the corresponding relative gains on NDIRIS-0405 are  $92.30\%$, $81.81\%$ and $85.71\%$. On CASIA.v4-Interval, the relative gains are $88.37\%$, $88.88\%$ and $68.75\%$. For IITD the corresponding relative gains are  $99.77\%$, $99.89\%$ and $98.33\%$. Also, from DET curves of Figure \ref{fig:Deep_DET} it can be clearly seen that our proposed method outperforms all the competing models.
We consistently outperformed our previous work, ICIP'20 across all datasets. This is attributed to a more texture aware encoder initialization (as previously shown in Figure \ref{fig:IITD_texture_reconstruct}) coupled with our pairwise learning (as opposed to offline matching in ICIP'20). It is to be noted that our relative gains over DeepIrisNet are highest on IITD dataset which has the smallest train set. This strongly encourages leveraging stagewise training and pairwise matching framework for data constrained datasets such as iris. Finally, we observe that our stagewise training strategy is consistently a better option than finetuning a pre-trained VGG-16.

\begin{table}[]

\centering
\caption{ Benchmarking the performance of our proposed model against deep learning methods on \textbf{NDIRIS-0405} dataset under 
\textit{Within Dataset} testing configuration.($\downarrow$) is better.}
\label{WDB_deep_Ndiris}
\begin{tabular}{|l|l|l|ll}
\cline{1-3}
Model                & EER (in\%) $\downarrow$ & AUC $\downarrow$ &  &  \\ \cline{1-3}
Finetuned VGG-16       & 7.87           & 0.026    &  &  \\
DeepIrisNet \cite{gangwar2016deepirisnet}   & 5.05          &  0.011   &  &  \\
ICIP'20 \cite{chakraborty2020unsupervised}          &  4.98          &   0.014  &  &  \\
Proposed   &  1.78        & 0.002     &  &  \\\cline{1-3}
\end{tabular}
\end{table}
\begin{table}[]
\centering
\caption{ Benchmarking the performance of our proposed model against deep learning methods on \textbf{CASIA.v4-Interval} dataset under 
\textit{Within Dataset} testing configuration.($\downarrow$) is better.}
\label{WDB_deep_Interval}
\begin{tabular}{|l|l|l|ll}
\cline{1-3}
Model                & EER (in\%) $\downarrow$ & AUC $\downarrow$ &  &  \\ \cline{1-3}
Finetuned VGG-16     & 11.08          &0.043   &  &  \\
DeepIrisNet \cite{gangwar2016deepirisnet}       & 10.65          &  0.045   &  &  \\
ICIP'20 \cite{chakraborty2020unsupervised}          &  6.11          &   0.016  &  &  \\
Proposed     &  3.33        & 0.005     &  &  \\\cline{1-3}
\end{tabular}
\end{table}
\begin{table}[]
\centering
\caption{ Benchmarking the performance of our proposed model against deep learning methods on \textbf{IITD} dataset under 
\textit{Within Dataset} testing configuration. ($\downarrow$) is better.}
\label{WDB_deep_IITD}
\begin{tabular}{|l|l|l|ll}
\cline{1-3}
Model                & EER (in\%)~$\downarrow$ & AUC~$\downarrow$ &  &  \\ \cline{1-3}
Finetuned VGG-16     & 11.2          & 0.0450   &  &  \\
DeepIrisNet \cite{gangwar2016deepirisnet}       & 7.92          &  0.0920   &  &  \\
ICIP'20\cite{chakraborty2020unsupervised}          &  3.84          &   0.0060  &  &  \\
Proposed     &  0.85        & 0.0001    &  &  \\\cline{1-3}
\end{tabular}
\end{table}




\textbf{Cross Dataset Performance:} 
We evaluate the performance of various competing deep-learning methods under this setting. As the traditional frameworks do not have a learning stage, this setting is not applicable to those methods.

This testing configuration is particularly challenging because the distribution of test images' statistics can be quite different from those of training images. A good performance under this setting is a sign of data efficient and generalised feature learning because a pre-trained system trained on one dataset is directly applied to the test dataset without any further fine-tuning on the test set. Thus, performance under this setting speaks of the generalization capability of different models. Also, this configuration mimics a product deployment scenario.

\begin{figure}[!t]
    \centering
    \includegraphics[scale=0.3]{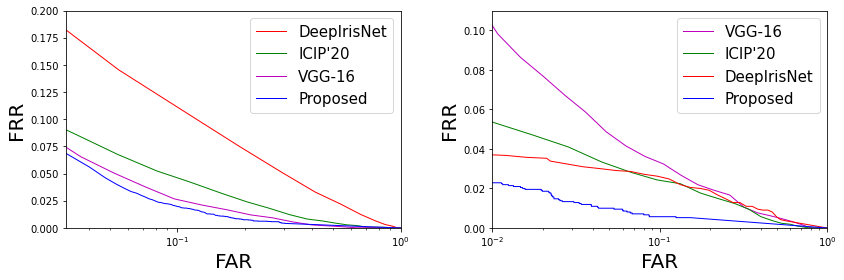}
    \caption{Comparative DET curves of proposed and competing deep learning baselines under challenging \textit{Cross Dataset} matching configuration. All the competing models are trained on NDIRIS-0405 dataset, and tested directly on \textbf{Left:} CASIA.v4-Interval, \textbf{Right:} IITD dataset without any futher fine-tuning. The curve closest to the bottom is desirable for better recognition performance, which is the case for our proposed model.}
    \label{fig:Deep_Cross_DET}
    
\end{figure}

For our study, we have trained all the models on NDIRIS-0405 dataset and tested directly on the test set of CASIA.v4-Interval and IITD datasets without any further fine tuning on the test data. The corresponding results are reported in Tables \ref{Cross_deep_IITD} and \ref{cross_deep_interval} and the respective DET curves for both the datasets are shown in Figure \ref{fig:Deep_Cross_DET}. As evident from both the tables as well as the DET curves, even under such challenging scenario our proposed model constantly outperforms the recent deep learning approaches. This experiment thus bolsters the data efficient feature learning, robustness and generalization capability of our proposed architecture.
\begin{table}[]
\centering
\caption{Comparing performance of our proposed model against recent deep learning techniques under 
\textit{Cross Dataset} testing configuration. All the models are trained on \textbf{NDIRIS-0405} and tested directly on \textbf{IITD} dataset without any further fine tuning. ($\downarrow$ is better.)}
\label{Cross_deep_IITD}
\begin{tabular}{|l|l|l|ll}
\cline{1-3}
Model    & EER (in \%)~$\downarrow$ & AUC~$\downarrow$ &  &  \\ \cline{1-3}
Finetuned VGG-16    &   4.80          & 0.013    &  &  \\
ICIP'20\cite{chakraborty2020unsupervised} & 3.66            &  0.008   &  &  \\
DeepIrisNet \cite{gangwar2016deepirisnet}    &   3.30          & 0.010    &  &  \\
Proposed  &   1.95          &  0.001   &  &  \\ \cline{1-3}
\end{tabular}
\end{table}


\begin{table}[]
\centering
\caption{Comparing performance of our proposed model against recent deep learning techniques under 
\textit{Cross Dataset} testing configuration. All the models are trained on \textbf{NDIRIS-0405} and tested directly on \textbf{CASIA.v4-Interval} without any further fine tuning. ($\downarrow$) is better.}
\label{cross_deep_interval}
\label{CB_Interval}
\begin{tabular}{|l|l|l|ll}
\cline{1-3}
Model    & EER (in \%)~$\downarrow$ & AUC~$\downarrow$ &  &  \\ \cline{1-3}
DeepIrisNet \cite{gangwar2016deepirisnet}    &   10.74         & 0.045    &  &  \\
Finetuned VGG-16    &   5.10          & 0.020    &  &  \\
ICIP'20\cite{chakraborty2020unsupervised} & 6.08           &  0.026   &  &  \\
Proposed &   4.68          &  0.019   &  &  \\ \cline{1-3}
\end{tabular}
\end{table}


\textbf{Performance Trend Under Training Data Reduction:} 
Here we perform an additional study to see the effect of reduction of training data on final recognition performance. We compare our model against our previous work, ICIP'20, and also against our Stage-2 architecture but with no autoencoder pre-training for encoder. We term this model as, $CombNet_R$.
\par We have conducted this study on NDIRIS-0405 dataset under \textit{Within Dataset} configuration. 
\par
\begin{figure}[!t]
    \centering
    \includegraphics[scale=0.55]{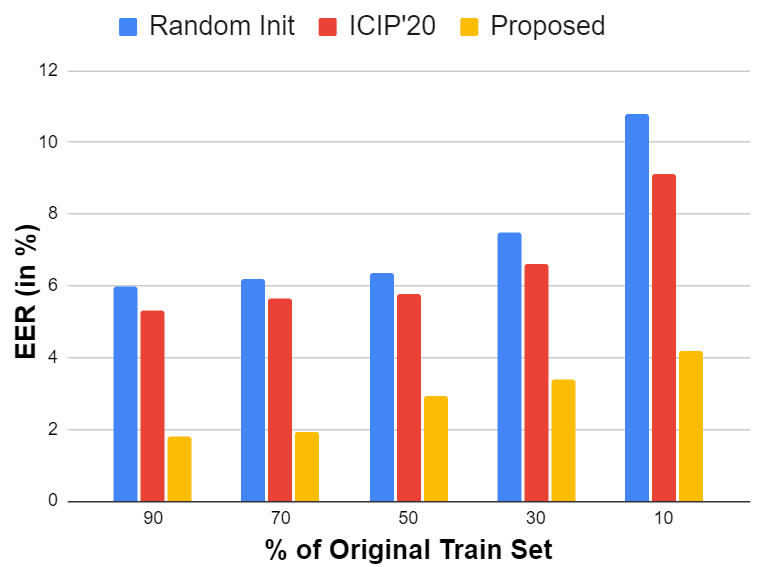}
    \caption{Performance of three different architectural variants under progressively reduced training size.~\textbf{Blue:} Randomly Initialised Encoder ($CombNet_R$). \textbf{Red:} Our previous work,  ICIP'20, \cite{chakraborty2020unsupervised} on stagewise learning where autoencoding (Stage-1) is optimised using only SSIM loss.~\textbf{Mustard:} Proposed pairwise learning where autoencoding (Stage-1) is optimised using SSIM + Relational + Denoising losses.} 
    \label{fig:training_reduction}

\end{figure}

 At 100\% i.e., under the original train split, we have 25images/subject on average for training. This enables us to experiment at as low as 10\% of original train data. Casia.v4-Interval and IITD datasets already have much less images per subject and so we exclude those datasets from this study. While reducing dataset size, we reduced the number of examples/class. For example, at 50\% setting, on average, there are only 12images/class for training. In Figure \ref{fig:training_reduction}, we visualize the EER metric of the three models as we gradually reduce the percentage of original train split to an extreme low of 10\%.
\par There are two important observations. First, the randomly initialized model consistently underperforms compared to both of the models in which the Stage-2 encoder is pre-trained in Stage-1. This systematically again bolsters the importance of stagewise training on small datasets. Second, the proposed model outperforms our previous ICIP'20 \cite{chakraborty2020unsupervised} at all levels of train data size-- this manifests the efficacy of our proposed loss functions and training strategies over our previous model.

\textbf{Comparing Network Complexities:} 
Iris recognition is becoming increasing popular in many mobile or handheld devices to grant security access. In such applications, lightweight models are preferred. In Table \ref{tab_param}, we show that on NDIRIS-0405 dataset, our proposed model utilizes only 0.91\%, 1.95\% and 77.01\% of total parameters as that used in DeepIrisNet, VGG-16 and ICIP'20. On CASIA.v4-Interval, the corresponding values are, 0.31\%, 0.66\% and 27.91\%. On IITD dataset the corresponding values are 0.31\%, 0.67\% and 28.34\%
Also, the respective total \textbf{FLO}ating \textbf{P}oint Operations (FLOPs) for our model is the least across all the competing methods. 
It is to be noted, that for the other models, we are considering FLOPs only till  feature tapping layer (feature extraction stage) of the respective network. We are excluding the additional computations required for the offline matching from this calculation. But, our model does not have such an offline matcher. So, in reality, FLOPs of the remaining three models will be even higher.
Low paramaters and FLOPs counts for our model means it will have less memory footprint and will require less computational resource during a recognition task.
It is to be noted, that reduction in computational cost compared to our previous work, ICIP'20 \cite{chakraborty2020unsupervised} is marginal. This is primarily because, in this current work, we focused mainly on better loss functions and training strategies with subtle modification in base architecture.


\begin{table}[]
\centering
\caption{Paramter count and FLOPs of competing deep networks for NDIRIS-0405, CASIA.v4-Interval  and IITD. In the table columns the respective datset names are mentioned as NDIRIS, CASIA and IITD.}
\label{tab_param}
\begin{tabular}{|l|c|c|c|c|c|c|}
\hline
\multicolumn{1}{|c|}{\textbf{Models}}                       & \multicolumn{3}{c|}{\textbf{\begin{tabular}[c]{@{}c@{}}\#Params\\ (in $10^6$)\end{tabular}}} & \multicolumn{3}{c|}{\textbf{\begin{tabular}[c]{@{}c@{}}FLOPs\\ (in $10^6$)\end{tabular}}} \\ \hline
                                                            & NDIRIS                   & CASIA                  & IITD                    & NDIRIS                  & CASIA                 & IITD                   \\ \cline{2-7} 
DeepIrisNet \cite{gangwar2016deepirisnet}  & 293.23                        & 292.35                             & 292.15                  & 7610.49                       & 7609.61                            & 7609.41                 \\
Finetuned VGG-16                                            & 137.17                         & 136.30                              & 136.09                   & 10166.47                         & 10165.59                             & 10165.39                   \\
ICIP'20 \cite{chakraborty2020unsupervised} & 3.48                          & 3.26                               & 3.21                    & 792.27                        & 792.05                             & 792.00                  \\
Proposed                                                    & 2.68                          & 0.91                               & 0.91                    & 207.52                        & 94.07                              & 94.07                   \\ \hline
\end{tabular}
\end{table}



\section{Conclusion}
In summary, this current work can be seen a data efficient paradigm of training lightweight deep networks on datasets which are rich in texture and also lacks multitudes of training samples for each subject.
This paper built upon on the stagewise learning strategy \cite{chakraborty2020unsupervised} which proved to be an effective training paradigm on iris datasets which have limited training data. Seeding from \cite{chakraborty2020unsupervised}, the paper presents optimization and architectural novelties which fosters better recognition performances for both \textit{Within-Dataset} and \textit{Cross-Dataset} configurations over a variety of iris datasets. Specifically, this paper presents a better autoencoding framework to train the Stage-1 encoder that acts as a feature initializer for the Stage-2 pairwise matching framework. In Stage-1, incorporation of relational data loss helped in learning better texture primitives while the denoising loss component aided in learning robust features. Combination of these two components in Stage-1 explicitly aided in Stage-2 matching performances
as can be seen from Table \ref{tab_ablation}. Moreover, our pairwise matching framework in Stage-2 further enhances recognition performance compared to that achieved from an usual offline matching paradigm. Our method consistently outperformed deep learning baselines which either exploits transfer learning or training an usual CNN network from scratch.
We also took a step further to show our model's efficacy under extreme data reduction scenario (Figure \ref{fig:training_reduction}), Finally, we showed our network offers significant savings in memory and FLOPs count compared to competing deep learning models. In summary, texture aware processing using data relation loss proved to enhance the recognition performance. Motivated by this, in future we wish to explore better texture reconstruction methodologies in Stage 1. Towards this end, an immediate extension of this work can be to explore the benefit of using perceptual loss \cite{ledig2017photo} which is commonly used for texture preserving image reconstruction tasks.


\bibliography{sn-bibliography}


\end{document}